\documentclass[conference]{IEEEtran}
\IEEEoverridecommandlockouts
\usepackage{cite}
\usepackage{amsmath,amssymb,amsfonts}
\usepackage{graphicx}
\usepackage{subfig}
\usepackage{textcomp}
\usepackage{xcolor}
\usepackage{inconsolata}
\usepackage{amsmath}
\usepackage{arydshln}
\usepackage{bbm}
\usepackage{hyperref}
\usepackage{fontawesome5}
\usepackage[linesnumbered,ruled,vlined]{algorithm2e}
\usepackage{amsmath} 
\usepackage{bm} 
\usepackage{multirow}
\usepackage{indentfirst}
\usepackage{booktabs}
\usepackage{caption}
\usepackage{adjustbox}
\def\BibTeX{{\rm B\kern-.05em{\sc i\kern-.025em b}\kern-.08em
    T\kern-.1667em\lower.7ex\hbox{E}\kern-.125emX}}
\begin{document}

\title{Dual-Teacher De-biasing Distillation Framework for Multi-domain Fake News Detection\\
\thanks{Jiayang Li and Xuan Feng have contributed to this work equally.}
\thanks{Tianlong Gu is the corresponding author.}
\thanks{This work was supported by the National Natural Science Foundation of China (Grant No. U22A2099).}
}

\author{\IEEEauthorblockN{1\textsuperscript{st} Jiayang Li}
\IEEEauthorblockA{\textit{Jinan University} \\
\textit{College of Cyber Security}\\
\textit{Engineering Research Center of}\\
\textit{Trustworthy AI (Ministry of Education)}\\
Guangzhou, China \\
ningljy@163.com}
\and
\IEEEauthorblockN{1\textsuperscript{st} Xuan Feng}
\IEEEauthorblockA{\textit{Jinan University} \\
\textit{College of Cyber Security}\\
\textit{Engineering Research Center of}\\
\textit{Trustworthy AI (Ministry of Education)}\\
Guangzhou, China \\
fenffef@163.com}
\and
\IEEEauthorblockN{2\textsuperscript{nd} Tianlong Gu}
\IEEEauthorblockA{\textit{Jinan University} \\
\textit{College of Cyber Security}\\
\textit{Engineering Research Center of}\\
\textit{Trustworthy AI (Ministry of Education)}\\
Guangzhou, China \\
gutianlong@jnu.edu.cn}
\and
\IEEEauthorblockN{3\textsuperscript{rd} Liang Chang}
\IEEEauthorblockA{\textit{Guilin University of Electronic Technology} \\
\textit{Guangxi Key Laboratory of Trusted Software}\\
Guilin, China \\
changl@guet.edu.cn}
}

\maketitle

\begin{abstract}
Multi-domain fake news detection aims to identify whether various news from different domains is real or fake and has become urgent and important. However, existing methods are dedicated to improving the overall performance of fake news detection, ignoring the fact that unbalanced data leads to disparate treatment for different domains, i.e., the domain bias problem. To solve this problem, we propose the Dual-Teacher De-biasing Distillation framework (DTDBD) to mitigate bias across different domains. Following the knowledge distillation methods, DTDBD adopts a teacher-student structure, where pre-trained large teachers instruct a student model. In particular, the DTDBD consists of an unbiased teacher and a clean teacher that jointly guide the student model in mitigating domain bias and maintaining performance. 
For the unbiased teacher, we introduce an adversarial de-biasing distillation loss to instruct the student model in learning unbiased domain knowledge. 
For the clean teacher, we design domain knowledge distillation loss, which effectively incentivizes the student model to focus on representing domain features while maintaining performance. Moreover, we present a momentum-based dynamic adjustment algorithm to trade off the effects of two teachers. Extensive experiments on Chinese and English datasets show that the proposed method substantially outperforms the state-of-the-art baseline methods in terms of bias metrics while guaranteeing competitive performance \footnote{Our codes are available at \url{https://github.com/ningljy/DTDBD}}. 
\end{abstract}

     
\section{Introduction}
With the popularity of social media platforms, multi-domain fake news detection has become urgent and important. The task aims to identify whether various news from different domains is real or fake \cite{nan2021mdfend, zhang2020fakedetector}. Despite the extraordinary achievements of extant methods in terms of average performance, they lack unbiased consideration of each domain \cite{nan2022improving}. Disparate treatment of fake news in different domains can lead to serious consequences \cite{wang2022streaming, truong2023fredom, xu2022can,zhu2022generalizing}. For example, misclassifying real disaster news as fake news potentially leads to delaying a user's emergency response to the danger. Therefore, reducing bias is one of the most critical aspects of deploying multi-domain fake news detection models in real-world scenarios.

\begin{table}[t]
\caption{Statistics on the amount of news in different domains (Weibo21 dataset \cite{nan2021mdfend}). \%Fake represents the percentage of fake news within each corresponding domain. \%News indicates the proportion of news articles in that domain as a percentage of the overall dataset.
(Ent. is an abbreviation for Entertainment)}
\begin{center}
\begin{tabular}{c|ccccc}
\hline
Domain & Science & Military & Education     & Disaster & Politics \\ \hline
\%Fake & 39.4    & 64.7     & 50.5          & 76.1     & 64.0     \\
\%News & 2.6     & 3.8      & 5.3           & 8.5      & 9.3      \\ \hline
Domain & Health  & Finance  & Ent. & Society  & Average  \\ \hline
\%Fake & 51.5    & 27.4     & 30.5          & 55.1     & 51.0     \\
\%News & 11.0    & 14.5     & 15.8          & 29.2     & 11.1     \\ \hline
\end{tabular}
\label{table:1}
\end{center}
\end{table}

\begin{table*}[t]
\caption{Functional Comparison among the Fake News Detection Methods.}
\begin{center}
\begin{tabular}{c|ccccc}
\hline
Method                            & Single-domain & Multi-domain & Debiasing & Bias Type & Dataset                        \\ \hline
BiGRU\cite{ma2016detecting}                              & \faCheck                        &                                &                             &                             & Twitter, Weibo                          \\
StyleLSTM\cite{przybyla2020capturing}                    & \faCheck                        &                                &                             &                             & StyleLSTM                               \\
DualEmo\cite{zhang2021mining}                            & \faCheck                        &                                &                             &                             & RumourEval-19, Weibo-16, Weibo-20       \\
EANN \cite{wang2018eann}                                 &                                 & \faCheck                       &                             &                             & Twitter, Weibo                          \\
Diachronic Bias Mitigation \cite{murayama2021mitigation} & \faCheck                        &                                & \faCheck                    & Diachronic                  & MultiFC, Horne17, Celebrity, Constraint \\
EDDFN\cite{silva2021embracing}                           &                                 & \faCheck                       &                             &                             & PolitiFact, Gossipcop, CoAID            \\
MDFEND\cite{nan2021mdfend}                               &                                 & \faCheck                       &                             &                             & Weibo21                                 \\
ENDEF\cite{zhu2022generalizing}                          & \faCheck                        &                                & \faCheck                    & Entity                      & Weibo, GossipCop                        \\
M3FEND\cite{zhu2022memory}                               &                                 & \faCheck                       &                             &                             & Weibo21, Politifact, Gossipcop, COVID   \\
Our                                                      &                                 & \faCheck                       & \faCheck                    & Domain                      & Weibo21, Politifact, Gossipcop, COVID   \\ \hline
\end{tabular}%
\label{table:11}
\end{center}
\end{table*}

The amount of news in different domains differs greatly due to real-time events, social trends, public information needs, and other factors \cite{shu2017fake,shetiya2022fairness}. The existing dataset also suffers from the following limitations owing to this property (taking Weibo21 \cite{nan2021mdfend} as an example, whose statistics are shown in Table \ref{table:1})): 1) the amount of fake news pieces is unbalanced across domains (e.g., Social news accounts for 29.2\% of the total number of fake news pieces, while Science accounts for only 2.6\%); 2) the rate of fake news pieces is practically unequal across domains (e.g., only 27.4\% of Finance news and 30.5\% of Entertainment news are fake. Yet, this percentage is up to 76.1\% and 64.0\% on Disaster news and Politics news); 3) models trained on unbalanced datasets may discriminate in predicting news from different domains. In addition, imbalanced datasets can lead to models learning spurious correlations between news labels and domain labels, which may affect generalization and incur serious domain bias problems. Thus, tackling the domain bias problem exposed by multi-domain fake news detection in a supervised setting requires more attention.

Existing single-domain fake news detection methods \cite{ruchansky2017csi,ma2016detecting,kwon2013prominent} are dedicated to improving detection performance while ignoring domain features of news. This leads them to be vulnerable in learning or even amplifying domain bias based on unbalanced datasets. In turn, most multi-domain fake news detection efforts \cite{nan2021mdfend,zhu2022memory} incorporate domain knowledge to guide the representation of more comprehensive features. However, it is possible that models lacking unbiased designs may unintentionally establish spurious correlations between the domain and the authenticity of the news. 
In fact, a piece of news may be related to multiple domains but has limited relevance to others. As such, forcing models to learn invariant features of all domains will further deteriorate the domain bias. Nevertheless, although directly applying existing debiasing methods \cite{wang2018eann,silva2021embracing} can mitigate bias to some extent, it will cause serious performance degradation. Hence, the trade-off between the de-biasing effectiveness and the prediction performance in multi-domain fake news detection tasks is an imperative problem to be solved.

In these regards, we propose the Dual-Teacher De-biasing Distillation framework (DTDBD), which mitigates domain bias arising from the unbalanced distribution of domains and categories in real-world datasets. Following the knowledge distillation methods, DTDBD adopts a teacher-student structure, where pre-trained large teachers instruct a student model. In particular, the DTDBD consists of an unbiased teacher and a clean teacher that jointly guide the student model. For the unbiased teacher, inspired by feature distillation \cite{romero2014fitnets}, we design a novel optimization objective called adversarial de-biasing distillation that transfers unbiased distribution as knowledge. By treating the unbiased distribution as a soft label, we prevent the student model from being solely responsible for learning the intricacies of the unbiased distribution from the unbiased teacher model. For the clean teacher, we introduce a domain knowledge distillation loss to encourage the student model to flexibly focus on multiple relevant domains for each domain. Furthermore, we present a momentum-based dynamic adjustment algorithm, which dynamically adjusts the weights of the unbiased and the clean teachers. The trade-off between unbiased representation and news representation enables the DTDBD to mitigate bias while maintaining competitive performance.

To sum up, the contributions can be summarized in the followings:
\begin{itemize} 
  \item We propose a novel Dual-Teacher De-biasing Distillation framework (DTDBD) to mitigate the domain bias in multi-domain fake news detection. It consists of an unbiased teacher and a clean teacher with distinct specialties to jointly guide the student model.
  \item To reduce the domain bias of the student model, we design an adversarial de-biasing distillation for the unbiased teacher. This distillation method captures the correlation of samples in the intermediate layer and utilizes it as knowledge for de-biasing distillation.
  \item Considering the trade-off between the impact of the two teachers, we introduce a momentum-based dynamic adjustment algorithm, which is determined by the change in the performance and bias metrics of the student model.
  \item Extensive experimental results demonstrate the effectiveness of our method in simultaneously improving model performance and mitigating domain bias. Additionally, our method achieves state-of-the-art performance on Chinese datasets, highlighting its superiority in multi-domain fake news detection.
\end{itemize}

\section{Related Work}

\subsection{Multi-domain Fake news detection}
Fake news detection aims to identify news pieces as real or fake \cite{zhou2019fake}. Traditional methods \cite{ruchansky2017csi,ma2016detecting,kwon2013prominent} focus on a single domain of fake news detection. These methods can be broadly classified into content-based and social context-based fake news detection. Content-based information includes news text \cite{sheng2022zoom,ma2019detect}, images \cite{wang2021multimodal,khattar2019mvae}, styles \cite{castillo2011information,przybyla2020capturing}, and emotions \cite{giachanou2019leveraging,zhang2021mining,choudhry2022emotion}. Social context-based information depends on user profiles \cite{dou2021user,shu2018understanding}, propagation networks \cite{liu2018early,nguyen2020fang,silva2021propagation2vec}, and crowd feedback \cite{shu2019defend,ma2018detect}.

In real-world scenarios, fake news typically originates from multiple different domains, thus multi-domain fake news detection has received attention. Wang et al. \cite{wang2018eann} first recognized the impact of news event diversity on fake news detection. They introduced an event adversarial neural network (EANN) that learned event-invariant features. Silva et al. \cite{silva2021embracing} proposed a framework with domain-specific and cross-domain knowledge to detect the authenticity of news from different domains. Nan et al. \cite{nan2021mdfend} collected a Chinese multi-domain fake news detection dataset Weibo21 from Sina Weibo, which includes nine domains: science, military, education, disasters, politics, health, finance, entertainment, and society. They developed a learnable domain gate to aggregate features extracted by multiple experts. Zhu et al. \cite{zhu2022memory} further designed a Domain Memory Bank to discover potential domain labels of news for guiding the aggregated features of domain adapters. However, the above efforts merely focus on performance improvements and ignore the problem of biases that exist among domains. Only a few studies have paid attentions to the bias problem in fake news detection. Zhu et al. \cite{zhu2022generalizing} employed causal diagrams to eliminate the entity bias between entities and news in the inference process. To mitigate diachronic bias, Murayama T et al. \cite{murayama2021mitigation} substituted Wikidata for proper names, encompassing individuals and geographical locations. The functional comparison of fake news detection related works are shown in Table \ref{table:11}.

\begin{table*}[t]
\centering
\caption{False Negative Rates (FNR) and False Positive Rates (FPR) of four advanced models on four disequilibrium domains: disaster, politics, finance, and entertainment. (Ent. is an abbreviation for Entertainment)}
\small
\begin{tabular}{c|cc|cc|cc|cc}
\hline
\multirow{2}{*}{Model} & \multicolumn{2}{c|}{Disaster} & \multicolumn{2}{c|}{Politics} & \multicolumn{2}{c|}{Finance} & \multicolumn{2}{c}{Ent.} \\ \cline{2-9} 
       & FNR    & FPR    & FNR    & FPR    & FNR    & FPR    & FNR    & FPR    \\ \hline
EANN \cite{wang2018eann}  & 0.0738 & 0.1756 & 0.0420 & 0.2115 & 0.1644 & 0.0938 & 0.1310 & 0.0735 \\
EDDFN \cite{silva2021embracing} & 0.0656 & 0.2674 & 0.1261 & 0.1923 & 0.1918 & 0.0729 & 0.1429 & 0.0735 \\
MDFEND \cite{nan2021mdfend}& 0.0574 & 0.1471 & 0.0588 & 0.1713 & 0.1370 & 0.0573 & 0.1429 & 0.0441 \\
M$^3$FEND\cite{zhu2022memory} & 0.0410 & 0.2059 & 0.0420 & 0.2308 & 0.1370 & 0.0573 & 0.1429 & 0.0245 \\ \hline
\end{tabular}
\label{table:2}
\end{table*}

\subsection{Domain adversarial training}
Domain adversarial training \cite{rong2023goddag} aims to map data from the source domain and target domain with different distributions into the same feature space. Simultaneously, the method expects the distances of differently distributed data to be as close as possible in this space. Ganin et al. \cite{ganin2015unsupervised} introduced the domain adversarial neural network (DANN), which has been successfully applied to mitigate bias by utilizing domain adversarial training. Kashyap et al. \cite{kashyap2022towards} tailored a domain adversarial method to mitigate texture bias in machine learning models. Liu et al. \cite{liu2020mitigating} employed domain adversarial learning to address gender bias in neural dialogue generation. Chowdhury et al. \cite{chowdhury2021adversarial} introduced “adversarial scrubbers”, an adversarial learning framework that intends to remove contextual representations and mitigate bias caused by demographic correlations. Choi et al. \cite{choi2019can} integrated a domain adversarial module into an action recognition framework to alleviate scene bias arising from action scene co-occurrence.

\subsection{Fair knowledge distillation}
Knowledge distillation \cite{hinton2015distilling} is a model compression method to achieve model lightweight and performance improvement by transferring knowledge from a complex model (teacher model) to a simple model (student model). Previous work has focused on how to transfer knowledge \cite{hinton2015distilling,romero2014fitnets,park2019relational} and what kind of knowledge to transfer \cite{ji2021show,du2020agree} in order to improve the performance of the student model.

Recently, several studies have found that model compression can affect fairness, and there has been growing attention on the role of knowledge distillation in bias mitigation tasks \cite{ahn2022knowledge, xu2022can}. Chai et al. \cite{chai2022fairness} adopted knowledge distillation to mitigate group bias without relying on demographic information. Liu et al. \cite{liu2022kdcrec} proposed a novel generic knowledge distillation framework to address bias in recommender systems. In addition, knowledge distillation has been leveraged to handle bias in information retrieval \cite{zhu2022debias} and face recognition \cite{dhar2021distill,jung2021fair}. Thus, knowledge distillation is a promising way to mitigate domain bias.

\section{Preliminary}
In this section, we initially present the fundamental notations and preliminary concepts employed in this paper. Subsequently, we proceed to formalize the problem of domain bias.

\subsection{Notations and Definitions}
\textbf{Definition 1} (Fake News Detection \cite{shu2017fake}). Given a news dataset $N= \{P, Y\}$, where $P= \{p_1, p_2, ..., p_n\}$ is a set of news text and $Y= \{y_1, y_2, ..., y_n\}$ is a set of news label. Each news item $p$ is associated with binary labels $y= \{0,1\}$, where $0$ indicates the news is real and $1$ indicates it is fake. Fake news detection aims to detect the authenticity of news by training a model to find the association between the news text $P$ and the label $Y$.

\textbf{Definition 2} (Multi-domain Fake News Detection \cite{nan2021mdfend}). We represent a multi-domain news dataset as $N_M = \{P, D, Y\}$, where $P$ represent the news texts, $Y$ denote the news label, and $D= \{d_1, d_2, ..., d_n\}$ is the domain of the news. Each news $p$ is assigned a domain label $d$ and a news label $y$. Multi-domain fake news detection is characterized by the use of domain labels for capturing the characteristics of news in the same domain to detect fake news.

\textbf{Definition 3} (Domain Disparate Mistreatment \cite{purificato2022graph}). Domain disparate mistreatment is a metric of domain bias. Given a fake news detection model $h(\cdot)$ and a sample $(p, d, y)$, where $p$ is the news text, $d$ is the news domain, and $y$ is the news label. It is called unbiased if the rate of true positives or false positives of this classifier on any two different domains $i$, $j$ is equal, i.e.,
\begin{equation}
\mathcal{P}(h(p) \neq y | y=0, d=i) = \mathcal{P}(h(p) \neq y | y=0, d=j).
\end{equation}
\begin{equation}
\mathcal{P}(h(p) \neq y | y=1, d=i) = \mathcal{P}(h(p) \neq y | y=1, d=j).
\end{equation}


\textbf{Definition 4} (Knowledge Distillation \cite{hinton2015distilling}). Knowledge distillation aims to enable the student model to learn from the teacher model by minimizing the discrepancy between their logits. In this context, the teacher model is typically a complex and high-performing model, while the student model is a lightweight model. Given a pre-trained teacher model $T$ and an untrained student model $S$, we denote the logits of the news data $P$ from the teacher model and the student model as $logits_T$ (soft labels) and $logits_S$, respectively.  It is essential to highlight that logits are response-based knowledge. Furthermore, in our method, we leverage feature-based distillation\cite{romero2014fitnets}, where the intermediate layer outputs of the model serve as knowledge to be transferred.

\subsection{Problem Statement}
Given a multi-domain news dataset $N_M = \{P, D, Y\}$, our objective is to train a fake news detection model, which treats news from different domains unbiasedly. In our problem setting, we adopt domain disparate mistreatment to measure the false negative rates (FNR) and false positive rates (FPR) of the model. Given the FNR and the FPR of a pre-trained model on two different domains $(FNR_i, FPR_i)$, $(FNR_j, FPR_j)$. The domain bias constraint can be expressed as:

\begin{equation}
    \forall i, j \in D,FNR_i \approx FNR_j.
\end{equation}
\begin{equation}
    \forall i, j \in D,FPR_i \approx FPR_j.
\end{equation}

\section{Analysis of Domain bias }

\subsection{ Domain bias in existing methods }A thorough analysis was conducted on four imbalanced domains using four advanced fake news detection models, revealing a noteworthy concern about the domain bias problem. Table \ref{table:2} illustrates that the Disaster and Political domains display significantly elevated FPR compared to the average. This implies that the models have developed a tendency to label news from domains with a substantial prevalence of fake news as fake. On the contrary, the Financial and Entertainment domains, which consist of a greater proportion of real news, exhibit higher FNR. This indicates that the models are more inclined to classify news from these domains as real.

\begin{figure*}[tpb]
\centerline{\includegraphics[scale=0.85]{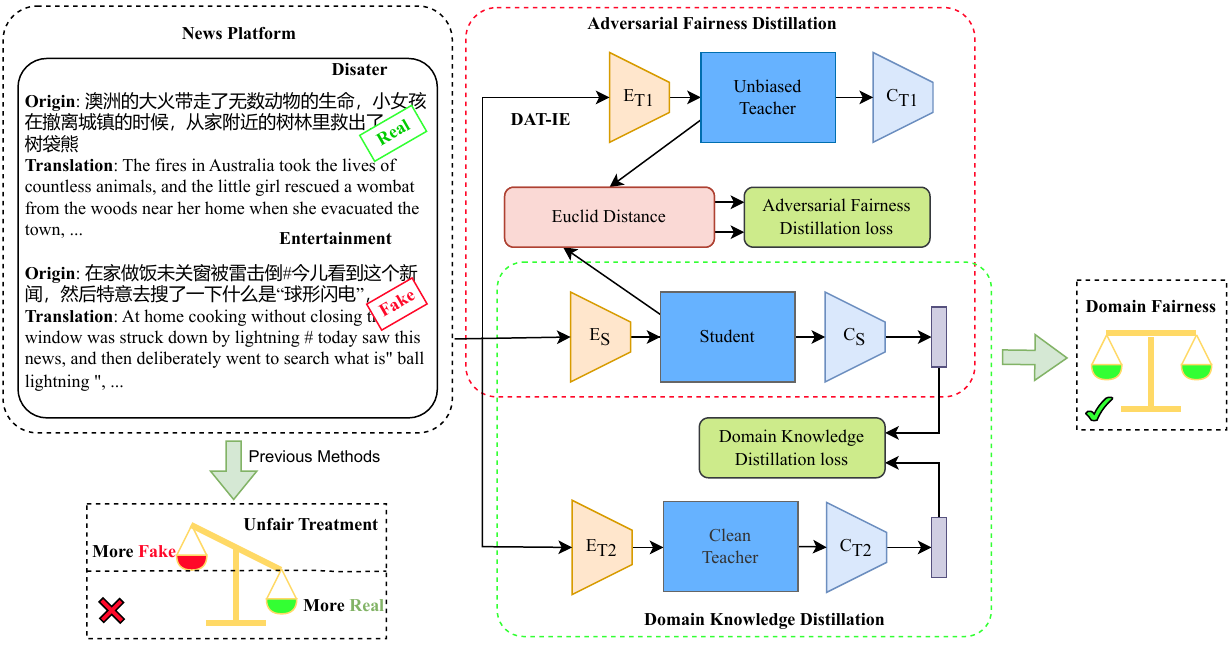}}
\caption{The framework of our DTDBD. In the process of DTDBD, We first train an unbiased teacher using the DAT-IE loss and select a fine-tuned multi-domain fake news detection model with a domain knowledge learning module as a clean teacher. Then, we calculate the adversarial de-biasing distillation loss and the domain knowledge distillation loss of the student through the guidance of the unbiased teacher and the clean teacher, respectively. Finally, we use a momentum-based dynamic adjustment algorithm to assign weights to the two losses and update the student model.}
\label{fig:1}
\end{figure*}

\subsection{ Challenges }
To tackle the domain bias problem, we emphasize the following two key points: 
\subsubsection{Integration of domain knowledge} The inclusion of domain knowledge in the learning process is crucial for mitigating domain drift and enhancing learning performance. In this context, domain drift refers to the fact that fake news originating from different domains exhibits significant variations in vocabulary, emotional tone, and writing style \cite{zhu2022memory}. However, it is essential to exercise caution when utilizing domain knowledge as a categorization factor, as it may inadvertently learn domain preferences, potentially impacting generalization and bringing the domain bias problem. 
\subsubsection{Consideration of fuzzy labels} The topics covered in real-world news are diverse, and a news item can often be relevant to multiple domains simultaneously \cite{zhu2022memory}. Therefore, the domain labels of news should be ambiguous; in other words, the domain labels should reflect the degree of similarity of the news to each domain. Incorporating fuzzy labels aids in identifying potentially relevant news beyond the confines of a single domain. However, excessive emphasis on data with low relevance may impede the acquisition of cross-domain knowledge. 

Initially, our objective was to eliminate spurious correlations between domains and news veracity. Nevertheless, we discovered that news from different domains can exhibit high correlations. Eliminating this correlation weakens the connection between news text and its veracity. Consequently, we adopt a dual-teacher knowledge distillation framework. This framework not only eliminates spurious correlations between domains and news veracity but also preserves correlations between news content and its truthfulness. By employing knowledge distillation, we ensure that the learned associations are more accurate and aligned with the actual truthfulness of the news.

\section{Framework}
In this section, we present a detailed explanation of the proposed DTDBD framework, which consists of adversarial de-biasing distillation and domain knowledge distillation. Additionally, we introduce a momentum-based dynamic adjustment algorithm utilized to adaptively adjust the weights of these two distillation methods. The overall framework of DTDBD is shown in Figure \ref{fig:1}.

\subsection{The DTDBD Framework}
To reduce domain bias while mitigating performance degradation in the context of multi-domain fake news detection, we propose a novel de-biasing framework called Dual-Teacher De-biasing Distillation (DTDBD). Unlike previous multi-teacher distillation methods \cite{furlanello2018born,yuan2021reinforced}, our method simultaneously focuses on two distinct metrics: bias and performance.

While some prior work has utilized adversarial knowledge distillation \cite{zhao2022enhanced,micaelli2019zero} to enhance the robustness of the student model by introducing adversarial examples for transferring knowledge from teacher models, robustness primarily remains a performance metric in anomalous situations. In contrast, DTDBD integrates two key distillation approaches: adversarial de-biasing distillation and domain knowledge distillation, leveraging an unbiased teacher and a clean teacher to guide the student model, respectively. It is worth noting that the weights of the unbiased and clean teachers are frozen during the distillation process.

The adversarial de-biasing distillation is designed to mitigate domain bias by using the unbiased teacher to transfer the knowledge related to unbiased distributions to the student model. On the other hand, the domain knowledge distillation leverages the clean teacher to transfer domain-specific knowledge to the student model, enabling it to effectively handle the nuances and characteristics of different domains.

By combining these two distillation approaches, DTDBD provides a comprehensive framework that guarantees performance while addressing domain bias in multi-domain fake news detection, resulting in a more balanced and effective framework. First, we select a fine-tuned multi-domain fake news detection model with a domain knowledge learning module as a clean teacher and train an unbiased teacher model. Second, the adversarial de-biasing distillation loss and the domain knowledge distillation loss of the student model are computed under the guidance of the unbiased teacher and the clean teacher, respectively. Finally, a momentum-based dynamic adjustment algorithm assigns weights to these two losses and subsequently updates the student model.

\subsection{Adversarial de-biasing distillation}
While enforcing the model to learn invariant features can mitigate bias to some extent, it can have a serious impact on performance. Therefore, we design the adversarial de-biasing distillation to reduce domain bias by transferring unbiased distribution as knowledge to the student model.
Considering that bias is reflected in the relative relationship among samples, we use the correlation between the intermediate features as unbiased distribution knowledge, and adopt Euclidean distance to measure the correlation of two samples. Given a set of intermediate features $F=\{f_k\}_{k=1} ^L $, which is obtained by a feature extractor, the correlation matrix \textbf{\textit{M}} can be expressed as:

\begin{equation}
    \boldsymbol{M}_{ij} = {\Vert f_a - f_b \Vert}_{2}^{2}.
\end{equation}

Then we perform adversarial de-biasing distillation in the intermediate layer, which enables the student model to learn correlations between samples mastered by the unbiased teacher. Adversarial de-biasing distillation can be formulated as:

\begin{equation}
\begin{split}
\mathrm{L}_{\mathrm{ADD}} =\tau^2\mathrm{KL}(&{LogSoftmax}(\boldsymbol{M_T})/\tau, \\
&{Softmax}(\boldsymbol{M_S})/\tau),    
\end{split}
\label{ADD}
\end{equation}
where KL is Kullback-Leibler divergence loss, $\tau$ is a temperature hyperparameter used in knowledge distillation. $M_T$ and $M_S$ follow from unbiased teacher $T_f (p|\theta_T)$ and student  $S (p|\theta_S)$, which represents student network $S$ with parameters $\theta_S$ and unbiased teacher network $T_f$ with parameters $\theta_T$.

For the design of the unbiased teacher, due to the large gap between the encoders of the different models and the difficulty of transferring the learned knowledge of unbiased distribution to each other, we set the structure of the unbiased teacher to be the same as that of the student model. Domain adversarial training was applied to get an unbiased teacher. Specifically, encoder $G_f (p;\Theta_f)$, domain classifier $G_d (f;\Theta_d)$, and label classifier $G_y (f;\Theta_y)$ can be described as the following optimization process:

\begin{equation}
\theta_f = \underset{\theta_f}{argmin} E(\theta_f, \theta_y, \theta_d).
\end{equation}

\begin{equation}
\theta_y = \underset{\theta_y}{argmin}E(\theta_f, \theta_y, \theta_d).
\end{equation}

\begin{equation}
\theta_d = \underset{\theta_d}{argmin}E(\theta_f, \theta_y, \theta_d).
\end{equation}

However, we find that on multi-domain fake news detection tasks, this approach causes the model to learn a shortcut, i.e., to learn only the common features of the domains most relevant to its own domain, while ignoring other relevant domains.

To solve this problem, we propose the domain adversarial training - information entropy (DAT-IE) loss, which encourages the model to focus on more relevant domains by adding information entropy loss to domain adversarial training. Information entropy can express the degree of uncertainty of a vector, which is the same as our goal in domain adversarial. Despite this may lead to the model failing to differentiate between domains instead of misclassification, Information entropy loss can be defined as:

\begin{equation}
    \mathrm{L}_{\mathrm{IE}} = G_d(f) \cdot log(G_d(f)^T).
\end{equation}

At this point, the DAT-IE loss can be expressed as:

\begin{equation}
\begin{split}
\mathrm{L}_{\mathrm{DAT-IE}} =&\mathrm{L}_{\mathrm{CE}}(G_y(f), y) +
\alpha\cdot\mathrm{L}_{\mathrm{CE}}(G_d(f), d) \\&+ \beta\cdot \mathrm{L}_{\mathrm{IE}}, \end{split}
\label{dat}
\end{equation}
where $L_{CE}$ denotes loss of cross entropy. In this paper, we set $\beta=0.2\alpha$ to extend the range of attention to invariant features.

\subsection{Domain knowledge distillation}
Previous work in multi-domain fake news detection \cite{zhu2022memory,liang2022fudfend} has demonstrated that fuzzy domain labels facilitate the accuracy of fake news detection. Moreover, we find that some news pieces from different domains may exhibit more similarities than those within the same domain.

Therefore, we design the domain knowledge distillation to encourage models to learn transferable unbiased domain knowledge which guarantees performance in bias mitigation. In domain knowledge distillation, we use a state-of-the-art fine-tuned multi-domain fake news detection model M$^3$FEND \cite{zhu2022memory} as a teacher for this part since it learns effectively shared domain knowledge by building domain knowledge learning modules. The loss of domain knowledge distillation process can be represented as:

\begin{equation}
\begin{split}
\mathrm{L}_{\mathrm{DKD}} =\tau^2\mathrm{KL}(&{LogSoftmax}({C_T(f_T)})/\tau, \\&{Softmax}({C_S(f_S)})/\tau),
\end{split}
\label{dkd}
\end{equation}
where $C_T(\cdot)$ and  $C_S(\cdot)$ denote the classifiers of the clean teacher and the student $S (p|\theta_S)$ respectively. $f_T$ and $f_S$ are intermediate features of clean teacher and student, respectively.

\subsection{Momentum-based dynamic adjustment algorithm}
In order to trade off the impact of the unbiased teacher and the clean teacher on the student model and prevent a single teacher from overplaying their role, we introduce a momentum-based dynamic adjustment algorithm. DTDBD uses the weighting of adversarial de-biasing distillation loss, domain knowledge distillation loss, and classification loss as the overall loss, which can be expressed as follows:

\begin{equation}
\begin{split}
\mathrm{L}_{\mathrm{overall}} (r) = &\omega_{\mathrm{ADD}} (r) \cdot \mathrm{L}_{\mathrm{ADD}} + \omega_{\mathrm{DKD}} (r)\cdot \mathrm{L}_{\mathrm{DKD}} \\& + \omega_{\mathrm{S}} (r)\cdot \mathrm{L}_{\mathrm{CE}},    
\end{split}
\label{overall}
\end{equation}
where $\omega_{\mathrm{ADD}}$, $\omega_{\mathrm{DKD}}$, $\omega_{\mathrm{S}}$ denote the weights of adversarial de-biasing distillation loss, domain knowledge distillation loss, and student classification loss, respectively.
We calculate the change in performance $\Delta F1$ and in bias metrics $\Delta Bias$ since the second epoch. The update process of $\omega_{ADD}$ and $\omega_{DKD}$ can be formulated as follows:

\begin{equation}
\begin{split}
\omega_{\mathrm{ADD}} (r) = &m \cdot \omega_{\mathrm{ADD}} (r-1) - \\&(1-m) \cdot (\Delta Bias - \Delta F1).
\end{split}
\label{wafd}
\end{equation}

\begin{equation}
\begin{split}
\omega_{\mathrm{DKD}} (r) =1 - \omega_{\mathrm{ADD}} (r),
\end{split}
\label{wdkd}
\end{equation}
where $m \in [0,1)$ is a momentum coefficient. 

\begin{algorithm}[t]
\SetAlgoNlRelativeSize{-2}
\SetNlSty{textbf}{\hspace*{-2em}}{}
\SetAlgoNlRelativeSize{+0}
\SetAlgoLined
\caption{DTDBD}
\label{alg:1}
\textbf{Input:} $P$: set of news, $R$: number of epochs;

\textbf{Output:} $D$: set of domain labels, $Y$: set of news labels;

Initialize student model $S (p|\theta_S)$, pre-trained clean teacher model $T_p$, and unbiased teacher model $T_f (p|\theta_T)$\;

\For{$r=0$ \textbf{to} $R-1$}{
  Compute $L_{\mathrm{DAT-IE}}$ loss using \eqref{dat}\;
  $\theta_T \leftarrow \theta_T - \eta \nabla_{\theta_T} L_{\mathrm{DAT-IE}}$\;
}

\For{$r=0$ \textbf{to} $R-1$}{
  Compute $L_{\mathrm{ADD}}$ loss using \eqref{ADD}\;
  Compute $L_{\mathrm{DKD}}$ loss using \eqref{dkd}\;
  
  \If{$r > 0$}{
    Compute $\omega_{\mathrm{ADD}}(r)$ and $\omega_{\mathrm{DKD}}(r)$ using \eqref{wafd} and \eqref{wdkd} respectively\;
  }
  Compute $L_{\mathrm{overall}}(r)$ using \eqref{overall}
  $\theta_S \leftarrow \theta_S - \eta \nabla_{\theta_S} L_{\mathrm{overall}}$\;
}
\end{algorithm}

We detail the DTDBD framework in Algorithm \ref{alg:1}.

\begin{table}[t]
\caption{Data Statistics of Chinese Dataset.}
\begin{center}
\begin{tabular}{cccccc}
\hline
Domain & Science & Military & Education  & Disaster & Politics \\ \hline
Fake   & 93      & 222      & 248   & 591      & 546      \\
Real   & 143     & 121      & 243   & 185      & 306      \\
Total  & 236     & 343      & 491   & 776      & 852      \\ \hline
Domain & Health  & Finance  & Ent.  & Society  & All      \\ \hline
Fake   & 515     & 362      & 440   & 1,471    & 4,488    \\
Real   & 485     & 959      & 1,000 & 1,198    & 4,640    \\
Total  & 1,000   & 1,321    & 1,440 & 2,669    & 9,128    \\ \hline
\end{tabular}
\label{table:3}
\end{center}
\end{table}

\begin{table}[t!]
\caption{Data Statistics of English Dataset.}
\begin{center}
\begin{tabular}{ccccc}
\hline
Domain & Gossipcop & Politifact & COVID & All    \\ \hline
Fake   & 5,067     & 379        & 1,317  & 6,763  \\
Real   & 16,804    & 447        & 4,750  & 22,001 \\
Total  & 21,871    & 826        & 6,067  & 28,764 \\ \hline
\end{tabular}
\label{table:4}
\end{center}
\end{table}

\section{Experiments}
\subsection{Experimental Settings}
\subsubsection{Datasets}We evaluate our DTDBD on Chinese dataset Weibo21\cite{nan2021mdfend} and English datasets including FakeNewsNet \cite{shu2020fakenewsnet} and COVID \cite{li2020mm}. The statistics of Chinese and English datasets are shown in Table \ref{table:3} and \ref{table:4}.

\begin{itemize}
  \item \textbf{Chinese Dataset}. Weibo21 \cite{nan2021mdfend} is a Chinese multi-domain fake news detection dataset designed to evaluate the average performance of fake news detection models across various domains. It consists of the news collected from Sina Weibo and is categorized into nine domains, including science, military, education, disaster, politics, health, finance, entertainment, and society.
  \item \textbf{English Dataset}. Following \cite{zhu2022memory,nan2022improving}, the English dataset we use is a merger of FakeNewsNet \cite{shu2020fakenewsnet} and COVID \cite{li2020mm}, including three domains: gossipcop, politics, and COVID.
\end{itemize}

\subsubsection{Baseline Models} In our study, we employed a state-of-the-art multi-domain fake news detection method \cite{zhu2022memory} and adopted the same baselines as used in that work. Our baseline models encompass the following 11 approaches:

\begin{table*}[htbp]
\caption{Performance and bias metrics comparison of DTDBD and baseline methods on Chinese datasets. MD and M$^3$ denote the use of MDFEND and M$^3$FEND as clean teachers in our method. The best and second-best results are highlighted in \textbf{bold} and \underline{underline}.}
\begin{center}
\resizebox{\textwidth}{!}{
\setlength{\tabcolsep}{2mm}
\begin{tabular}{c|ccccccccc|cccc}
\hline
\multirow{2}{*}{Methods} &
  \multirow{2}{*}{Science} &
  \multirow{2}{*}{Military} &
  \multirow{2}{*}{Education} &
  \multirow{2}{*}{Disaster} &
  \multirow{2}{*}{Politics} &
  \multirow{2}{*}{Health} &
  \multirow{2}{*}{Finance} &
  \multirow{2}{*}{Ent.} &
  \multirow{2}{*}{Society} &
  \multicolumn{4}{c}{Overall} \\ \cline{11-14} 
   &  &  &  &  &  &  &  &  &  & F1 & FNED & FPED & Total \\ \hline
BiGRU\cite{ma2016detecting} & 0.7479& 0.9129 & 0.8182 & 0.8574	& 0.8694	& 0.8850	& 0.8582	& 0.8671	& 0.8602 &	0.8754	& 0.6361	& 0.3538	& 0.9899\\

TextCNN\cite{kim2014convolutional} & 0.7890	& 0.9270	& \underline{0.9091}	& 0.8765	& 0.8680	& 0.9250	& 0.8935	& 0.8404	& 0.8801	& 0.8934	& 0.5730	& 0.4535	& 1.0265 \\

BERT  & 0.7566	& 0.9129	& 0.8887	& 0.8895	& 0.8728	& 0.9200	&0.8823 &	0.9076 &	0.8901	& 0.9032	& 0.6241	& 0.4160	 & 1.0401 \\

RoBERTa\cite{liu2019roberta} & 0.7166	& 0.9417	& 0.8382	& 0.8684	& 0.8806	& 0.9100	& 0.8000	& 0.8750 &	0.8754	& 0.8884	& 0.8287	& 0.4108	& 1.2395 \\

StyleLSTM\cite{przybyla2020capturing}  &  0.8196	& 0.9273	& 0.8582	& 0.8790	& 0.8697	& 0.9250	& 0.8932 &	0.9121	& 0.8881	& 0.9049	& 0.5616	& 0.5464	& 1.1080 \\

DualEmo\cite{zhang2021mining}  & 0.8271	& \underline{0.9419}	& 0.8384	& 0.8821	& 0.8665	& 0.9050	& 0.8889	&0.9133 & 0.8992	& 0.9027	& 0.5429	& 0.4261	& 0.9690 \\

EANN\cite{wang2018eann}  & 0.8487	& \underline{0.9419}	& 0.8485	& 0.8760	& 0.8525	& 0.9300	& 0.8833	& 0.8967 &	0.8963	& 0.9021	& 0.4438	& \underline{0.3410}	& 0.7848 \\

EANN\_NoDAT  & 0.8196	& 0.9417	& 0.9089	& 0.8684	& 0.8934	& 0.9350	& 0.9012	& 0.9069	& 0.9033	& 0.9132	& 0.5696	& 0.3964	& 0.9660 \\

MMoE\cite{ma2018modeling}  &  0.8594	& 0.9275	& 0.8888	& 0.8496	& 0.8618	& 0.9299	& 0.8417	& 0.8840	 & 0.8739	& 0.8911	& 0.4728	& 0.3982	& 0.8710 \\

MoSE &  0.8125	& 0.8384	& 0.8586	& 0.8014	& 0.8240	& 0.9050	& 0.8573	& 0.9108 &	0.8676	& 0.8825	& 0.5093	& 0.8786	& 1.3879 \\

EDDFN\cite{silva2021embracing} & 0.8271	& 0.9273	& 0.8574	& 0.8765	& 0.8244	& 0.9374	& 0.8573	& 0.8802 &	0.8749	& 0.8912	& 0.5790	& 0.5597	& 1.1387 \\

EDDFN\_NoDAT & 0.8021	& 0.9270	& 0.9088	& 0.8694	& 0.8525	& 0.9247	& 0.8414	& 0.8810	& 0.8624	& 0.8916	& 0.5507	& 0.4530	& 1.0037 \\

MDFEND\cite{nan2021mdfend} & 0.8487	& 0.9417	& 0.8917	& 0.8872	& 0.8597	& 0.9400	& \underline{0.9020}	& 0.9121 &	0.8903	& 0.9120	& 0.5795	& 0.5250	& 1.1045
       \\

M$^3$FEND\cite{zhu2022memory} &  0.8196	& 0.9417	& 0.8787	& 0.8684	& 0.8856	& \underline{0.9450}	& 0.9016	& \textbf{0.9318}	& \underline{0.9131}	& 0.9207	& 0.5867	& 0.5099	& 1.0966
    \\
\hdashline
Our(MD) & \underline{0.9030} 	& \underline{0.9419}	& 0.8987	& \textbf{0.9060}	& \textbf{0.9152}	& \underline{0.9450}	& 0.8893	& 0.9023	& \underline{0.9131} & 	\underline{0.9213}	& \underline{0.4345}	& \textbf{0.3155}	& \underline{0.7500} \\

Our(M$^3$) & \textbf{0.9259}	& \textbf{0.9546}	& \textbf{0.9292}	&  \underline{0.8955}	&  \underline{0.9087}	& \textbf{0.9500}	& \textbf{0.9074} & 	\underline{0.9293}	& \textbf{0.9234}	& \textbf{0.9290}	& \textbf{0.3446}	& 0.4038	& \textbf{0.7484} \\
\hline      
\end{tabular}}
\label{table:5}
\end{center}
\end{table*}

\begin{itemize}
  \item Roberta \cite{liu2019roberta}. This is a pre-trained model that employs dynamic masking language modeling techniques. We used it as an encoder with frozen parameters to preserve learned representations. The model incorporates a Multilayer Perceptron (MLP) for classification.
  \item BiGRU \cite{ma2016detecting}. It is a text encoder based on a common recurrent neural network (RNN) architecture. In our implementation, we utilized a one-layer BiGRU with a hidden layer size of 300.
  \item TextCNN \cite{kim2014convolutional}. Specifically designed for processing and analyzing text data, TextCNN employs a convolutional neural network architecture. We utilized five convolutional kernels with different kernel steps of 1, 2, 3, 5, and 10, each with 64 channels.
  \item StyleLSTM \cite{przybyla2020capturing}. This model utilizes a bidirectional LSTM as an encoder. The encoded features are then fed into the MLP along with style features to obtain prediction results.
  \item DualEmo \cite{zhang2021mining}. It employs BiGRU as an encoder, and sentiment features are input to the MLP along with the text features. The encoder settings are similar to the BiGRU method.
  \item MMoE \cite{ma2018modeling}. This model adopts MLP as an expert network and combines representations from multiple experts using a gating mechanism.
  \item MoSE. Similar to MMoE, MoSE replaces the MLP with an LSTM as an expert network.
  \item EANN \cite{wang2018eann}. It consists of a feature extractor, a fake news classifier, and an event discriminator. Its objective is to identify and remove event-specific features while retaining shared features among different events. We utilized the setup described in \cite{zhu2022memory}.
  \item EDDFN \cite{silva2021embracing}. This framework aims to retain domain-specific and cross-domain knowledge in news data for detecting fake news across different domains. We followed the setup described in \cite{nan2021mdfend}.
  \item MDFEND \cite{nan2021mdfend}. It is a multi-domain detection model that employs learnable domain gates to aggregate expert networks. TextCNN serves as an expert network, and the setup of each expert network is similar to the baseline TextCNN approach.
  \item M$^3$FEND \cite{zhu2022memory}. It is a state-of-the-art multi-domain fake news detection model that uses domain adapters to aggregate semantics, sentiment, and style to obtain a comprehensive news representation while using a domain memory bank to generate latent domain labels for the sample.
\end{itemize}

\begin{table*}[t]
\caption{Performance and bias metrics comparison of DTDBD and baseline methods on English datasets. MD and M$^3$ denote the use of MDFEND and M$^3$FEND as clean teachers in our method. The best and second-best results are highlighted in \textbf{bold} and \underline{underline}.}
\begin{center}
\resizebox{0.7\textwidth}{!}{
\setlength{\tabcolsep}{2mm}
\begin{tabular}{c|ccc|cccc}
\hline
\multirow{2}{*}{Method} & \multirow{2}{*}{Gossipcop} & \multirow{2}{*}{Politics} & \multirow{2}{*}{COVID} & \multicolumn{4}{c}{Overall}       \\ \cline{5-8} 
                        &                          &                         &                        & F1     & FNED   & FPED   & Total  \\ \hline
BiGRU\cite{ma2016detecting}                   & 0.7774                   & 0.7726                  & 0.9016                 & 0.8048 & 0.2125 & 0.1317 & 0.3442 \\
TextCNN\cite{kim2014convolutional}                 & 0.7953                   & 0.7145                  & 0.8851                 & 0.8114 & \underline{0.1619} & 0.1173 & 0.2792 \\
RoBERTa\cite{liu2019roberta}                & 0.7998                   & 0.8033                  & 0.9131                 & 0.8258 & 0.2918 & 0.2315 & 0.5233 \\
StyleLSTM\cite{przybyla2020capturing}               & 0.7983                   & 0.8123                  & 0.9264                 & 0.8268 & 0.2031 & 0.0767 & 0.2798 \\
DualEmo\cite{zhang2021mining}                 & 0.7941                   & 0.8204                  & 0.9007                 & 0.8194 & 0.2480 & 0.1282 & 0.3762 \\
EANN\cite{wang2018eann}                    & 0.7934                   & 0.7731                  & 0.8749                 & 0.8019 & \textbf{0.1196} & 0.1475 & \underline{0.2671}\\
EANN\_NoDAT             & 0.7857                   & 0.7386                  & 0.8734                 & 0.8062 & 0.3381 & 0.1304 & 0.4685 \\
MMoE\cite{ma2018modeling}                    & 0.8047                   & 0.8486                  & \textbf{0.9453}                 & 0.8380 & 0.3079 & \underline{0.0750} & 0.3829 \\
MoSE                    & 0.7982                   & \underline{0.8553}                  & 0.9380                 & 0.8324 & 0.3723 & 0.1156 & 0.4879 \\
EDDFN\cite{silva2021embracing}                   & 0.7862                   & \textbf{0.8605}                  & 0.9307                 & 0.8217 & 0.3495 & 0.0754 & 0.4249 \\
EDDFN\_NoDAT            & 0.7986                   & 0.8451                  & \underline{0.9423}                & 0.8343 & 0.4156 & 0.0859 & 0.5015 \\
MDFEND\cite{nan2021mdfend}                  & \underline{0.8080}                   & 0.8473                  & 0.9331                 & \underline{0.8433} & 0.4376 & 0.1076 & 0.5452 \\
M$^3$FEND \cite{zhu2022memory}                  & \textbf{0.8237}                   & 0.8478                  & 0.9392               & \textbf{0.8454} & 0.4397 & 0.1472 & 0.5869 \\
\hdashline
Our(MD)         & 0.8025                   & 0.8005                  & 0.9259                 & 0.8294 & 0.1779 & 0.0830 & \textbf{0.2609} \\
Our(M$^3$)         & 0.8073                   & 0.8291                  & 0.9332                 & 0.8359 & 0.2021 & \textbf{0.0677} & 0.2698 \\ \hline

\end{tabular}}
\label{table:6}
\end{center}
\end{table*}

In addition, we evaluated the versions of EANN and EDDFN that do not contain the domain countermeasure module, which are denoted as EANN\_NoDAT and EDDFN\_NoDAT.

It is important to highlight that among these baseline models, only EANN, EDDFN, MDFEND, and M$^3$FEND incorporate the domain labels as part of their input.

\subsubsection{Evaluation Metrics} Consistent with many previous studies \cite{liu2021authors,park2018reducing,lertvittayakumjorn2020find,liu2019incorporating}, we utilize false positive equality differences (FPED) and false negative equality differences (FNED) \cite{dixon2018measuring} to measure bias, along with F1 scores to assess performance. FPED quantifies the absolute difference between the overall FNR and the FPR for each domain, while FNED performs a similar calculation for false negatives, as shown in \eqref{FPED} and \eqref{FNED}. We use Total to denote the sum of FNED and FPED.
\begin{equation}
\begin{split}
FPED = \sum_{d \in D} |FPR - FPR_d|.
\end{split}
\label{FPED}
\end{equation}

\begin{equation}
\begin{split}
FNED = \sum_{d \in D} |FNR - FNR_d|.
\end{split}
\label{FNED}
\end{equation}

\subsubsection{Teacher and Student Networks} For the adversarial de-biasing distillation, we employ a network with the same architecture as the student model for domain adversarial training. This allows us to obtain an unbiased teacher model that facilitates the migration of the knowledge of unbiased distribution. In the domain knowledge distillation stage, we select two state-of-the-art multi-domain fake news detection methods, namely MDFEND \cite{nan2021mdfend} and M$^3$FEND \cite{zhu2022memory}, as the clean teacher models due to their effectiveness in capturing domain knowledge. It is worth noting that MDFEND and M$^3$FEND have 8.14M and 11.36M trainable parameters, respectively.  To ensure comparability, we design a student network called TextCNN-S, which utilizes commonly used network structures and had 7.71M trainable parameters. TextCNN-S incorporate BERT with the activation of layer 11 and employ five convolutional kernels with 64 channels, each with different kernel steps of 1, 2, 3, and 5 as the encoder. MLP was used for classification purposes.

\subsubsection{Training Setup} Our framework is implemented by PyTorch and trained on an NVIDIA GeForce RTX 3090. In accordance with the experimental setup outlined in \cite{zhu2022memory}, we measure the bias metrics of the baseline models and implement our approach. All baseline models were trained following the settings in the original paper and \cite{zhu2022memory}. For the teacher-student distillation framework proposed in this paper, we freeze the weights of both teachers during the distillation process. In addition, all the experiments were conducted on all domains. Our student model, TextCNN-U, uses a learning rate of 0.001 during domain adversarial training. In the case of our DTDBD framework, we set the initial learning rate to 0.0001, and the weights of the two teachers were dynamically adjusted using \eqref{wafd} and \eqref{wdkd}.

\subsection{Performance Comparison}
In this section, we evaluate the effectiveness of DTDBD on both Chinese and English multi-domain fake news detection datasets. The evaluation results are presented in Tables \ref{table:5} and \ref{table:6}, which show the F1 scores for each domain and overall metrics including F1, FPED, and FNED. The best-performing results are highlighted in bold, while the second-best results are underlined.

\begin{table*}[t]
\caption{Performance and bias metrics on the ablation study of our DTDBD on the Chinese dataset. DND and ADD are abbreviations of domain knowledge distillation and adversarial de-biasing distillation. Student+DAT-IE denotes the use of our improved domain adversarial training on the student model. Student+DND means using a clean teacher (M$^3$FEND) to guide the student in domain knowledge distillation. Student+ADD refers to the usage of only adversarial de-biasing distillation. Student+DTDBD indicates our DTDBD method. DAA denotes the momentum-based dynamic adjustment algorithm. The best and second-best results are highlighted in \textbf{bold} and \underline{underline}.}
\begin{center}
\resizebox{0.8\textwidth}{!}{
\setlength{\tabcolsep}{3mm}
\begin{tabular}{c|cccc|cccc}
\hline
\multirow{2}{*}{Model} & \multicolumn{4}{c|}{TextCNN-S}    & \multicolumn{4}{c}{BiGRU-S}       \\ \cline{2-9} 
                       & F1     & FNED   & FPED   & Total  & F1     & FNED   & FPED   & Total  \\ \hline
Student                & 0.9136 & 0.7161 & 0.4059 & 1.1220 & 0.8999 & 0.6503 & 0.4577 & 1.1080 \\
Student+DAT-IE         & 0.8967 & \textbf{0.3409} & \underline{0.3347} & \textbf{0.6756} & 0.8743 & {0.4735} & \textbf{0.3229} & \underline{0.7964} \\
Teacher(M$^3$)            & \underline{0.9207} & 0.5867 & 0.5099 & 1.0996 & 0.\textbf{9207} & 0.5867 & 0.5099 & 1.0996 \\ \hline
Student+DND            & 0.9185 & 0.6721 & 0.4271 & 1.0992 & 0.9127 & 0.5272 & 0.4418 & 0.9690 \\
Student+ADD            & 0.9109 & 0.4741 & \textbf{0.3053} & 0.7794 & 0.8980 & 0.5294 & \underline{0.3391} & 0.8685 \\
w/o DAA            & 0.9213 & 0.4908 & 0.4597 & 0.9505 & 0.9131 & \underline{0.4361} & {0.4253} & 0.8614 \\
Our(M$^3$)           & \textbf{0.9290} & \underline{0.3446} & {0.4038} & \underline{0.7484} & \underline{0.9142} & \textbf{0.4138} & 0.3491 & \textbf{0.7629} 
\\
\hline
\end{tabular}}
\label{table:7}
\end{center}
\end{table*}

\subsubsection{Domain unfairness measurement} By analyzing the results of the baseline methods, we observe that most single-domain and multi-domain fake news detection methods exhibit high FPED and FNED values. This indicates that these methods tend to learn domain bias from unbalanced datasets. We also find that EANN achieves the best debiasing results among the baseline models, suggesting that enforcing the model to learn cross-domain features is beneficial in mitigating domain bias. However, the performance of EANN is significantly lower than that of EANN\_NODAT, which validates the effectiveness of domain adversarial training. The reason behind this discrepancy is that the adversarial structure ignores scenarios where a news item may be related to multiple domains with varying degrees of relevance. Notably, EDDFN\_NODAT demonstrates a significant performance improvement compared to EDDFN on the English dataset, but the change is less pronounced on the Chinese dataset. This may be attributed to the fact that EDDFN incorporates an intra-domain knowledge learning module, which reduces the influence of the domain adversarial module in learning cross-domain features. Furthermore, since the English dataset only consists of three domains with substantial content gaps between them, the intra-domain knowledge learning module does not have a noticeable impact.

\begin{table*}[t]
\caption{Comparison of traditional DAT and our DAT-IE in performance and bias metrics. Student+DAT indicates the use of traditional domain adversarial training. Student+DAT-IE denotes the use of our improved domain adversarial training on the student model. The best and second-best results are highlighted in \textbf{bold} and \underline{underline}.}
\begin{center}
\resizebox{0.8\textwidth}{!}{
\setlength{\tabcolsep}{3mm}
\begin{tabular}{c|cccc|cccc}
\hline
\multirow{2}{*}{Model} & \multicolumn{4}{c|}{TextCNN-S}    & \multicolumn{4}{c}{BiGRU-S}       \\ \cline{2-9} 
                       & F1     & FNED   & FPED   & Total  & F1     & FNED   & FPED   & Total  \\ \hline
Student                & \textbf{0.9136} & 0.7161 & 0.4059 & 1.1220 & \textbf{0.8999} & 0.6503 & 0.4577 & 1.1080 \\
Student+DAT            & 0.8856 & \underline{0.3719} & \underline{0.3807} & \underline{0.7526} & 0.8599 & \underline{0.4836} & \underline{0.3290} & \underline{0.8126} \\
Student+DAT-IE         & \underline{0.8967} & \textbf{0.3409} & \textbf{0.3347} & \textbf{0.6756} & \underline{0.8743} & \textbf{0.4735} & \textbf{0.3229} & \textbf{0.7964} \\ \hline
\end{tabular}}
\label{table:8}
\end{center}
\end{table*}

\subsubsection{Results for DTDBD} Regarding our proposed method, TextCNN-U trained with DTDBD achieves state-of-the-art debiasing results while also attaining state-of-the-art performance on the Chinese datasets. This demonstrates that DTDBD is an effective approach that not only reduces domain bias but also mitigates the performance degradation problem. The main reason behind this success lies in the dual-teacher distillation structure employed by DTDBD, which transfers both the knowledge of unbiased distribution and domain knowledge to the student model. However, on the English dataset, the performance of our method is slightly lower than that of MDFEND and M$^3$FEND, but there is a significant improvement in domain bias mitigating. This discrepancy can be attributed to the fact that the English dataset comprises only three domains with substantial differences in news content, making it challenging to learn important cross-domain knowledge.

Additionally, we find that M$^3$FEND is more effective in imparting domain knowledge compared to MDFEND. While both methods employ a soft sharing mechanism to learn cross-domain knowledge, M$^3$FEND constructs a memory matrix for each domain, allowing for a richer representation of domain information by calculating the similarity of news samples to the memory matrix to obtain the domain label distribution.

\begin{figure*}[ht]
\centerline{\includegraphics[scale=0.4]{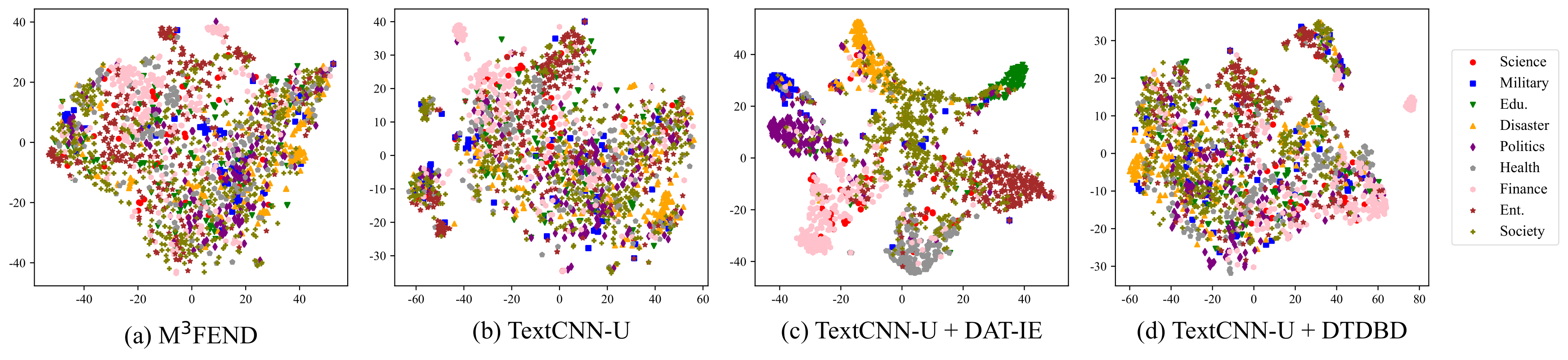}}
\caption{Visualization of intermediate features using the t-SNE on Chinese dataset. M$^3$FEND is a state-of-the-art method for multi-domain fake news detection and also the clean teacher in our domain knowledge distillation. TextCNN-U is the student model we use. +DAT-IE indicates training students with our domain adversarial training and also the teacher in the adversarial de-biasing distillation. + DTDBD indicates training the student model using our method.}
\label{fig:2}
\end{figure*}

\subsection{Ablation Study}
\subsubsection{Importance of different components} To gain a deeper understanding of the teaching characteristics of the two teachers in our DTDBD, the role of information entropy loss, and the effects of momentum-based dynamic adjustment
algorithm, we conduct a series of ablation studies. We employ the TextCNN-U model as the student model. Student+DND and Student+ADD indicate training the student model using domain knowledge distillation and adversarial de-biasing distillation, respectively. Student+DTDBD represents our DTDBD method. DAA denotes the momentum-based dynamic adjustment algorithm. Additionally, we introduce a BiGRU-S model to demonstrate the effectiveness of each distillation module, which utilizes a frozen BERT and a one-layer BiGRU with a hidden size of 300 for feature extraction, followed by an MLP for classification.

\subsubsection{Effectiveness of adversarial de-biasing distillation} As demonstrated in Table \ref{table:7}, adversarial de-biasing distillation significantly mitigates the domain bias of the model, indicating the effectiveness of using unbiased distribution as knowledge. Furthermore, we find that compared to domain adversarial training, adversarial de-biasing distillation mitigates the problem of performance degradation while bias mitigation is improved. This aligns with our objective of not excessively forcing the model to focus on learning invariant features but rather learning them in domains with a high degree of relevance whenever possible.

\subsubsection{Effectiveness of domain knowledge distillation} As shown in Table \ref{table:7}, we discover that domain knowledge distillation can improve the performance of the student model. The reason is that the clean teacher transfers effective domain knowledge to the students, which is not available in the student model. We also see that domain knowledge distillation slightly reduces the domain bias of the model, suggesting that the knowledge distillation method itself can limit the learning of redundant knowledge.

\begin{figure*}[t]
\centerline{\includegraphics[scale=0.9]{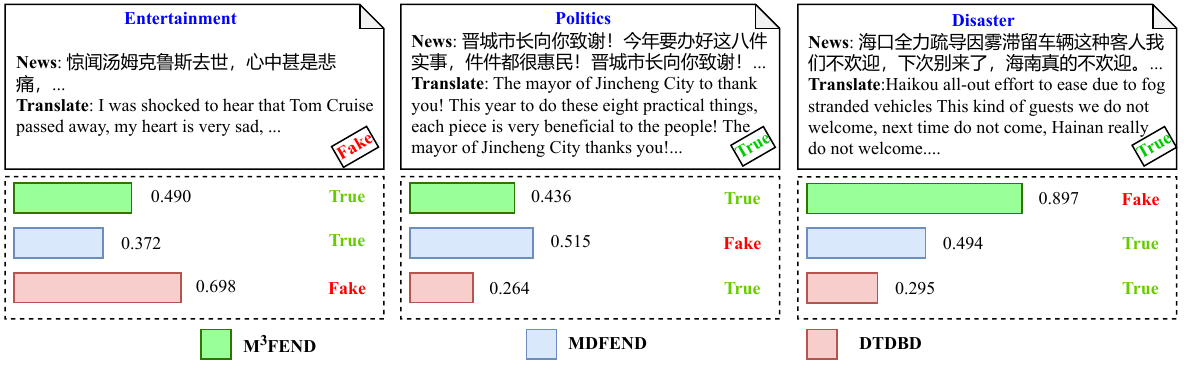}}
\caption{Case studies of three news pieces. The upper part of the figure shows the three news pieces and the lower part presents the probabilistic prediction results for M$^3$FEND, MDFEND, and our DTDBD, respectively.}
\label{fig:3}
\end{figure*}

\subsubsection{Effectiveness of momentum-based dynamic adjustment algorithm} As shown in Table \ref{table:7}, we observe that enhancements in both performance and fairness can occur even in the absence of the momentum-based dynamic adjustment algorithm, underscoring the significance of addressing fairness distillation and domain knowledge distillation. Nevertheless, it is worth noting that the improvements in fairness tend to be relatively modest while achieving further performance enhancements poses greater challenges. This can be attributed to the absence of an adversarial module in both the student model and the clean teacher model. Consequently, the student model tends to predominantly acquire knowledge from the clean teacher, somewhat neglecting the contributions of the unbiased teacher. This results in an insufficient acquisition of fairness-related knowledge during the learning process, ultimately limiting the potential for further performance improvement.

\subsubsection{Effectiveness of information entropy loss} We evaluate the impact of information entropy loss on two student models. Student+DAT indicates the use of traditional domain adversarial training, while Student+DAT-IE refers to our modified domain adversarial training incorporating information entropy loss. As shown in Table \ref{table:8}, we observe that DAT-IE helps further mitigate the domain bias and improve performance compared to the DAT approach. This is because the DAT approach allows the model to learn a shortcut by focusing only on the most relevant domains, ignoring the extraction of shared knowledge from other domains. On the contrary, the information entropy loss encourages the model to prioritize features common across all domains, highlighting the importance of shared features in reducing domain bias.

\subsection{Visualization of DTDBD} 
To visually illustrate the effectiveness of our method in learning cross-domain features, we employ dimensionality reduction using t-SNE to project the intermediate features of the test data into a 2-dimensional space, as shown in Figure \ref{fig:2}. It is important to maintain a nuanced perspective on the sample distribution, considering the correlation between news from different domains.

We observe that in the M$^3$FEND and TextCNN-U models, there are multiple areas containing samples from only one or a few domains. This suggests that these models may learn redundant domain-specific knowledge or focus primarily on shared features from a limited number of domains.

In the TextCNN-U+DAT-IE model, the aforementioned situation is more pronounced. This can be attributed to the incorporation of a DAT-IE loss in the news classification training, which compels the model to concentrate on the most relevant domains. Consequently, the model exhibits a higher degree of separation between domains in the feature space. 

In comparison, the TextCNN-U model trained using our DTDBD method exhibits more diverse patterns in domain representations. In 2(d), most regions contain samples from multiple domains, demonstrating that DTDBD effectively mitigates the phenomenon of domain spurious correlation. Importantly, DTDBD not only avoids equal focus on each domain but also exhibits the ability to precisely prioritize domains with high relevance. These visual representations provide insights into how our method enables the model to learn shared features across domains while accurately capturing the relevance of individual domains.

\subsection{Case Study} 

The case studies of the three news pieces are shown in Figure \ref{fig:3}. From Case 1, it can be seen that when traditional fake news detectors encounter domains where fake news predominates (e.g., entertainment and finance), they are more likely to misclassify this news as fake news. Conversely, in domains where real content is abundant, such as disasters and politics, previous methods suffer from identical domain bias (as shown in Case 2).  In contrast, our DTDBD framework effectively mitigates the influence of domain information on model decisions by utilizing domain knowledge distillation and adversarial de-biasing distillation.

Furthermore, it can be seen from Cases 1, 2, and 3 that the dual-teacher framework DTDBD can predict news labels more accurately and with higher prediction confidence. This reflects the role of the clean teacher in the model performance. On the other hand, it also indicates the effectiveness of the momentum-based dynamic adjustment algorithm for the trade-off between debiasing and detection performance.

\section{Conclusion}
In this paper, we propose the Dual-Teacher De-biasing Distillation framework (DTDBD), which mitigates domain bias arising from the unbalanced distribution of real-world datasets. DTDBD consists of an unbiased teacher and a clean teacher that jointly guide the student model in reducing domain bias and mitigating the performance degradation problem. For the unbiased teacher, we design an adversarial de-biasing distillation loss that incorporates unbiased distribution as a form of knowledge. For the clean teacher, we introduce a domain knowledge distillation loss to incentivize the student model to focus on learning the knowledge across correlated domains. Furthermore, we present a momentum-based dynamic adjustment algorithm to balance the effects of two teachers on the student model. We conduct extensive experiments on English and Chinese multi-domain fake news datasets to evaluate the effectiveness of DTDBD. The results demonstrate that DTDBD effectively mitigates domain bias, improves performance, and achieves state-of-the-art performance on Chinese datasets.



\bibliographystyle{IEEEtran}
\bibliography{ieee}

\end{document}